\pdfoutput=1

\documentclass[11pt]{article}

\usepackage[preprint]{acl}

\usepackage{times}
\usepackage{latexsym}
\usepackage[utf8]{inputenc}
\usepackage[T1]{fontenc}
\usepackage{CJKutf8}
\usepackage[vietnamese, polish, estonian, finnish, main=english]{babel}
\usepackage{graphicx}
\usepackage{hyperref}
\usepackage{multirow} 
\usepackage{float} 
\usepackage{amsmath} 
\usepackage{adjustbox} 
\usepackage{verbatim} 
\usepackage{makecell}
\usepackage{xparse}

\ExplSyntaxOn
\NewDocumentCommand{\xhrulefill}{O{}}
 {
  \group_begin:
  \severin_xhrulefill:n { #1 }
  \group_end:
 }

\keys_define:nn { severin/xhrulefill }
 {
  height .dim_set:N    = \l_severin_xhrule_height_dim,
  thickness .dim_set:N = \l_severin_xhrule_thickness_dim,
  fill .skip_set:N     = \l_severin_xhrule_fill_skip,
  height .initial:n    = 0pt,
  thickness .initial:n = 0.4pt,
  fill .initial:n      = 0pt plus 1fill,
 }

\cs_new_protected:Nn \severin_xhrulefill:n
 {
  \keys_set:nn { severin/xhrulefill } { #1 }
  \leavevmode
  \leaders\hrule 
    height \dim_eval:n { \l_severin_xhrule_thickness_dim + \l_severin_xhrule_height_dim }
    depth  \dim_eval:n { -\l_severin_xhrule_height_dim }
  \skip_horizontal:N \l_severin_xhrule_fill_skip
  \kern 0pt
}
\ExplSyntaxOff

\usepackage[T1]{fontenc}

\usepackage[utf8]{inputenc}

\usepackage{microtype}

\title{GPTs Are Multilingual Annotators for Sequence Generation Tasks}

\author{Juhwan Choi\textsuperscript{1}, Eunju Lee\textsuperscript{2}, Kyohoon Jin\textsuperscript{2} \and Youngbin Kim\textsuperscript{1,2} \\\\
  \textsuperscript{1}Department of Artificial Intelligence, Chung-Ang University \\
  \textsuperscript{2}Graduate School of Advanced Imaging Sciences, Multimedia and Film, Chung-Ang University \\
  \texttt{\{gold5230,dmswn5829,fhzh123,ybkim85\}@cau.ac.kr} \\
}

\begin{document}
\maketitle
\begin{abstract}
Data annotation is an essential step for constructing new datasets. However, the conventional approach of data annotation through crowdsourcing is both time-consuming and expensive. In addition, the complexity of this process increases when dealing with low-resource languages owing to the difference in the language pool of crowdworkers. To address these issues, this study proposes an autonomous annotation method by utilizing large language models, which have been recently demonstrated to exhibit remarkable performance. Through our experiments, we demonstrate that the proposed method is not just cost-efficient but also applicable for low-resource language annotation. Additionally, we constructed an image captioning dataset using our approach and are committed to open this dataset for future study. We have opened our source code for reproducibility.\footnote{\url{https://github.com/c-juhwan/gpt-multilingual-annotator}}
\end{abstract}

\section{Introduction}
With the evolution of deep learning methods, various tasks in the NLP domain have demonstrated remarkable performance. However, training deep learning models requires a substantial amount of labeled data. Data annotation, a process of gathering unlabeled data and labeling them, plays a crucial role in fulfilling this data demand.

However, as the conventional procedure of data annotation is mainly conducted manually using human annotators, it cannot meet the growing demand for labeled data with an increase in the size of deep learning models \cite{qiu2020pre}. Moreover, it is significantly challenging to recruit annotators for low-resource languages \cite{pavlick-etal-2014-language}.

To address the lack of labeled data and improve the performance of the model, the concept of pre-trained language model (PLM) was introduced. These PLMs have been trained on a large amount of text corpus to acquire a general knowledge of languages \cite{radford2018improving, devlin-etal-2019-bert}. By fine-tuning these models to specific downstream task, it was able to achieve performance improvement without the need for additional labeled data.

With the evolution of PLMs via the enlargement of their sizes owing to increased training data, the development of a large language model (LLM) with massive parameter size enabled few-shot learning from the context of the given prompt \cite{brown2020language}. Accordingly, the diverse capabilities of LLMs have been investigated \cite{zhao2023survey}.

However, despite their impressive abilities and adaptability, these LLMs cannot be actively exploited for downstream tasks because of the cost constraints and demand for hardware resources caused by their extensive model size. Additionally, fine-tuning these models for specific purposes remains challenging due to their massive parameter size. Consequently, training models for downstream tasks through labeled data is still the dominant approach for practical applications \cite{yu2023open}.

Data annotation refers to the creation of labeled data by assigning gold labels to unlabeled data. Traditionally, data annotation was mainly conducted by human labelers using crowdsourcing platforms, such as Amazon mechanical turk (MTurk), and these platforms have aided the creation of modern, large-scale datasets. Recently, to address these limitations of crowdsourcing-based data annotation and achieve a cost-efficient means to collect labeled data, several studies have proposed the utilization of LLMs as alternative annotators in place of human labelers \cite{wang-etal-2021-want-reduce, ding-etal-2023-gpt, gilardi2023chatgpt, jiao2023chatgpt, li-etal-2023-coannotating, zhang-etal-2023-llmaaa, he2023annollm, bansal2023large}. These studies have shown the possibility of cost-efficient and automatic data annotation through LLMs, such as GPT-3.

\begin{figure*}[h]
    \centering
    \includegraphics[width=1\textwidth]{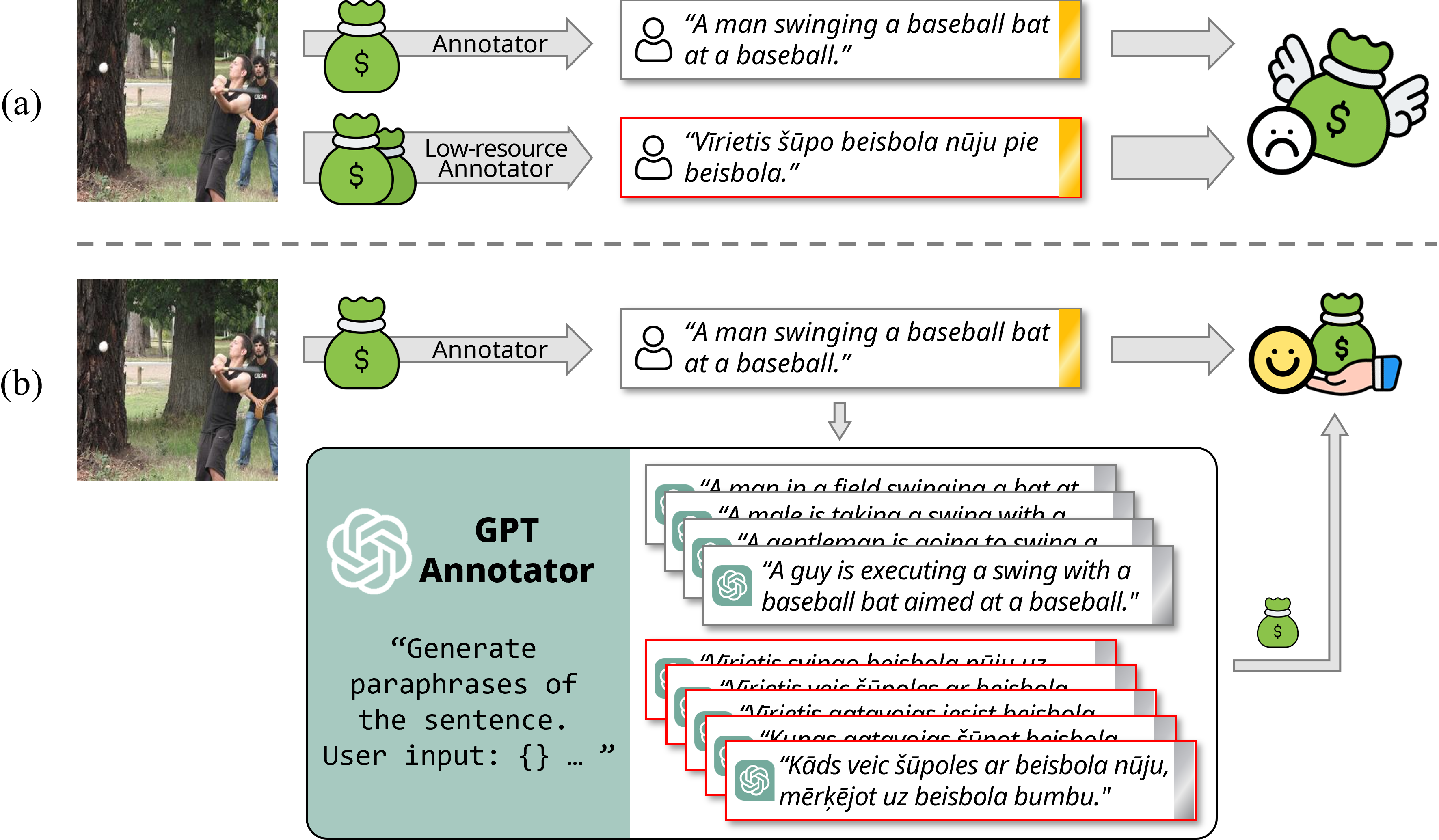}
    \caption{Overall concept of our GPT annotator. (a) Conventional annotation process for image captioning task, which is performed by multiple human annotators and expensive. Moreover, it is more expensive to hire human annotators for low-resource languages. (b) The annotation process of proposed GPT annotator. With one gold caption by a single human annotator, the GPT annotator automatically generates silver captions, as well as captions in other languages, resulting in a cost-efficient dataset construction.}
\label{fig-main}
\end{figure*}

However, as these existing studies mainly focused on simple tasks, such as text classification, additional investigation is required to apply these approaches to numerous subtasks of natural language processing. Moreover, the potential of automatic data annotation via LLMs has not been explored for languages other than English. As previously highlighted, projects in low-resource languages may suffer from the high cost of data annotation, necessitating the need for automatic annotators for languages beyond English.

In this study, we proposed a strategy that leverages LLMs as an assistant annotator to aid human annotators in image captioning task and text style transfer task. As depicted in Figure~\ref{fig-main}, the conventional process of establishing datasets for image captioning task required a considerable number of human annotators to generate five gold annotations for each image, resulting in a high cost for dataset construction in languages beyond English. Moreover, the quality of the annotated data varies depending on the proficiency of the human annotators \cite{rashtchian-etal-2010-collecting}. Similarly, the annotation process for text style transfer required significant human effort, including quality control \cite{rao-tetreault-2018-dear, briakou-etal-2021-ola}.

This study demonstrated the ability of LLMs to serve as assistant annotators for human annotators at a reasonable cost by generating multiple silver sentences for each gold annotation written by one single human annotator. Specifically, we proposed a cost-efficient process to construct multilingual language datasets by exploiting the GPT annotator. Particularly, we utilized GPT-4, which exhibits enhanced multilingual capabilities \cite{openai2023gpt4}, to autonomously produce diverse sentences in another language from a single English sentence, even if the human annotator is not familiar with the target language. Moreover, the cost of the GPT annotator is constant as the cost is determined by the length of the processed token, regardless of the language. This highlights the efficiency of the proposed GPT annotator as an annotation method for low-resource language, which is more expensive and time-consuming compared to English.

Employing this method, we developed an image captioning dataset in Latvian, Estonian, and Finnish — which are well-known low-resource languages — by employing the GPT annotator. In this scenario, a single human annotator, who lacks knowledge of the target language, provides one English gold caption for each image. Through the experiment, we demonstrated that the proposed method achieves better performance compared to machine translation method. We open these datasets to support future studies. Additionally, we release software to easily perform data annotation process described in this paper.

Our contributions are summarized as follows:
\begin{itemize}
\item To the best of our knowledge, this is the first work to explore the possibility of LLM as a multilingual annotator.
\item To the best of our knowledge, this is the first study to employ LLM as an automatic annotator for image captioning task and text style transfer task.
\item Our experiment reveals the ability of GPT annotators to serve as human annotators at a reasonable cost.
\item We release an annotation software to easily perform the method described in the paper, as well as three image captioning datasets in Latvian, Estonian, and Finnish.
\end{itemize}

\section{Related Work}
GPT-3 has demonstrated that LLMs can conduct in-context learning from few-shot prompts. Accordingly, various LLMs with different characteristics have been proposed \cite{zhao2023survey}. For example, based on the findings that LLMs can be further enhanced via human instruction and feedback \cite{ouyang2022training}, ChatGPT\footnote{\url{https://openai.com/blog/chatgpt}} and its backbone GPT-3.5 with various abilities have emerged \cite{leiter2023chatgpt, yang2023harnessing, liu2023summary}. In addition, the cutting-edge GPT-4 \cite{openai2023gpt4} is a progressed version of GPT-3.5, with a longer input sequence, improved multilingual ability, and image inception ability.

With the advancement of LLMs, studies have been conducted to augment given human-annotated data \cite{yoo2021gpt3mix, whitehouse-etal-2023-llm}, or to annotate unlabeled data and train models for downstream tasks. One of the early studies in this field \cite{wang-etal-2021-want-reduce} proposed an automatic annotation method that demonstrated the ability of GPT-3 to annotate a greater amount of data compared to human annotators at a lower labeling cost, resulting in higher performance at the same cost, and this strategy was observed to outperform GPT-3 itself. In addition, the study investigated the possibility of a collaboration between human and GPT annotators by leveraging the confidence of the automatic annotation of GPT to perform active labeling by human annotators.

Following this approach, subsequent studies expanded the annotation capabilities of GPT-3 to not just label unlabeled data but also create labeled data leveraging external knowledge, or even from scratch \cite{ding-etal-2023-gpt}. Meanwhile, a methodology was proposed to transfer the abilities of LLMs into a smaller model by generating a rationale for the labeled data, enhancing the performance of the small model \cite{hsieh2023distilling}.

With the emergence of ChatGPT, an improved version of GPT-3 that enables enhanced flexibility across diverse tasks, researchers have proposed its application for data annotation. ChatGPT has been reported to outperform crowdworkers in text classification tasks in certain cases with the same instructions \cite{gilardi2023chatgpt}. Additionally, studies observed that ChatGPT even surpassed expert labelers in the annotations of political texts \cite{tornberg2023chatgpt}. These results have led researchers to examine the annotation abilities of ChatGPT across various domains \cite{zhu2023can}.

Recent studies have expanded the application of LLMs as annotators, from language understanding tasks, such as text classification or inference, to text generation tasks. For example, a previous study reported improved performance in query-focused summarization by reducing the noise of ChatGPT \cite{laskar2023cqsumdp}. Additionally, dialogue generated by ChatGPT has been observed to demonstrate comparable quality to reference dialogues written by human annotators \cite{labruna2023unraveling}.

These studies indicate the capability of LLMs, including ChatGPT, to perform as an effective annotator for not just text understanding tasks but also text generation tasks, which are more complex and challenging to annotate. However, the application of these abilities of LLMs to various natural language processing tasks is still limited and underexplored. In this study, we proposed an LLM-based annotation method for image captioning task and text style transfer task, which has not been investigated in previous studies. Furthermore, we validated the feasibility of LLMs as an autonomous multilingual annotator, which has not been explored in previous works.

\section{Method}
\subsection{Task Formulation}

We first define a dataset $D$, which is composed of the data pair $d=(X, Y)$. In image captioning task, $X$ denotes a given image and $Y =\{y_{g_{1}}, y_{g_{2}}, ..., y_{g_{5}}\}$ is corresponding captions that describe $X$. In this paper, $g$ means ``gold'', which represents a human-annotated sentence. Similarly, in text style transfer task, $X$ denotes the original sentence and $Y_{g}$ indicates human-annotated pair sentence with desired style.

Traditionally, multiple human annotators are used to write descriptions for unannotated data $X$ to construct such datasets, especially for image captioning, which requires multiple captions for each image. However, as previously discussed, this entirely human-based annotating process is expensive and time-consuming. Our GPT annotator aims to construct a data pair by autonomously generating silver sentences and reduce the time and cost consumption of data annotation process.

Additionally, we explore the multilingual ability of the GPT annotator. The cost of data annotation varies by language. Especially, Low-resource languages are associated with higher cost and high time consumption for the collection of annotated data \cite{9733711, guemimi2021iterative, li2019santlr, kim2021commonsense}. This phenomenon is caused by the language pool of the crowdworkers \cite{pavlick-etal-2014-language} and the difficulty of training low-resource language natives \cite{lin-etal-2018-platforms}. In this study, we propose a method to employ the GPT annotator as a multilingual annotator through the adaptation of GPT-4, which has significantly improved multilingual ability \cite{openai2023gpt4}. 

\subsection{Assistant Multilingual Annotator for Image Captioning Task}

To achieve the aforementioned goal, we synthesized the given human-annotated caption $y_{g_{1}}$ by utilizing the GPT model, and generated a set of paraphrases $\{y_{s_{2}}, ..., y_{s_{5}}\}$ based on $y_{g_{1}}$.

We configured a well-designed prompt $P$, as the input for GPT to achieve this object. As it has been reported that LLMs perform significantly better with examples rather than zero-shot \cite{brown2020language}, the prompt $P$ includes an one-shot desired example. The process of generating sentences through GPT can be expressed as follows.

\begin{equation}
\label{eq1}
    \{y_{s_{2}},\;...,\;y_{s_{5}}\} = \textrm{GPT}(P,\;y_{g_{1}})
\end{equation}

The machine-annotated caption produced in Eq.~\ref{eq1} is used to construct a new data pair, $d'=(X, {y_{g_{1}}, y_{s_{2}}, ..., y_{s_{5}}})$, and a downstream task model is trained using dataset $D'$, a collection of these $d'$. Consequently, GPT can be used to assist human annotators with image captioning task.

In addition, to employ our GPT annotator as multilingual annotator, it first synthesizes a data pair with one single human annotation in English, $d^{\textit{src}}=(X, y_{g_{1}}^{\textit{eng}})$ to reduce the cost of hiring multiple human annotators. Secondly, the GPT annotator generates a set of paraphrases in a target language $\{y_{s_{1}}^{\textit{tgt}},..., y_{s_{5}}^{\textit{tgt}}\}$. This process is performed through a prompt $P^{\textit{tgt}}$ with information in the target language, including a one-shot desired example. We found it helpful to jointly generate English sentence and its translation rather than solely generate sentences in the target language, as English sentence guides the generation of target language sentence. Specific prompts can be found in Appendix~\ref{sec:appendix-prompt-cap-kor}. The described process can be expressed as follows.

\begin{equation}
\label{eq1}
    Y_{\textit{tgt}}=\{y_{s_{1}}^{\textit{tgt}},\;...,\;y_{s_{5}}^{\textit{tgt}}\} = \textrm{GPT}(P^{\textit{tgt}},\;y_{g_{1}}^{\textit{eng}})
\end{equation}

The dataset in target language $D^{\textit{tgt}}$ can be constructed through $d^{\textit{tgt}}=(X,Y^{\textit{tgt}})$ obtained by the GPT annotator, and a downstream task model in the target language can be trained using this $D^{\textit{tgt}}$. This overall process enables the construction of a dataset $D^{\textit{tgt}}$ in any designated language with only one single annotation in English by utilizing the LLM. Furthermore, this process is performed without any intervention of a human annotator who is fluent in the target language, reducing the cost of hiring expert annotators in the target language.

\begin{figure}[t]
    \centering
    \includegraphics[width=1\linewidth]{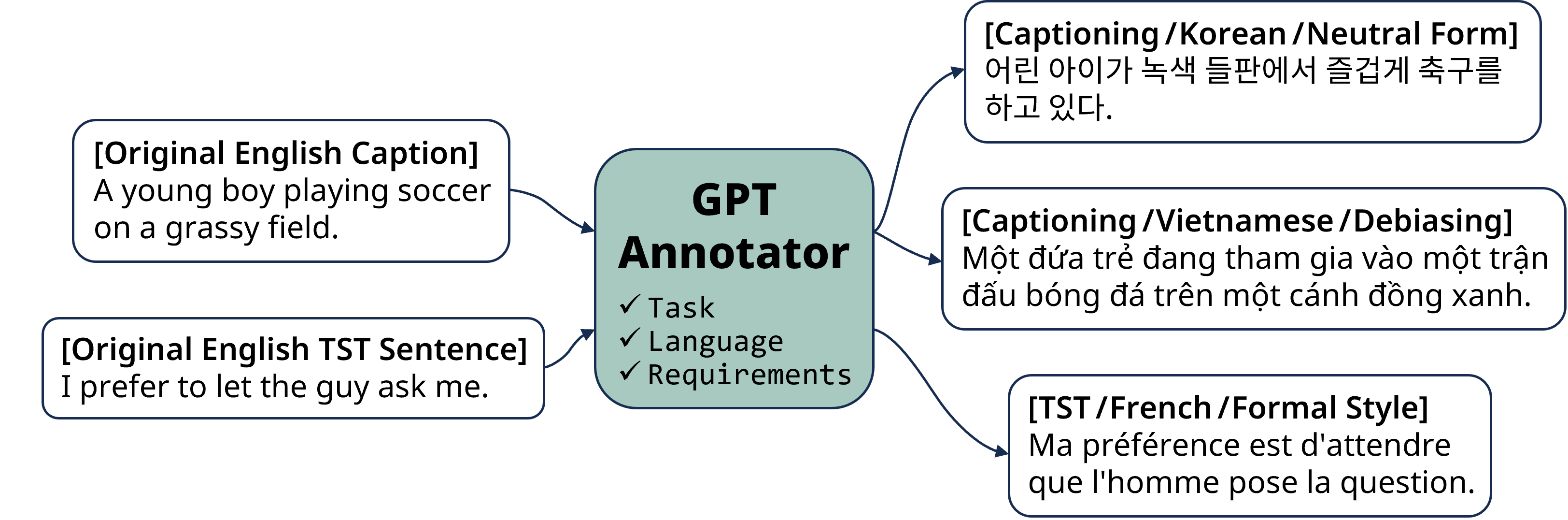}
    \caption{Our GPT annotator can generate various datasets with configurable prompts, primarily regarding task, language, and specific requirements.}
\label{fig-method}
\end{figure}

\subsection{Assistant Multilingual Annotator for Text Style Transfer Task}

For text style transfer task, we first analyze the given data pair $d^{\textit{src}}=(X^{\textit{eng}}, Y_{g}^{\textit{eng}})$ written in English through the GPT annotator. Nextly, the GPT annotator creates a translated version of the pair and its paraphrase in target language, $d_{1}^{\textit{tgt}}=(X_{s_{1}}^{\textit{tgt}}, Y_{s_{1}}^{\textit{tgt}})$ and $d_{2}^{\textit{tgt}}=(X_{s_{2}}^{\textit{tgt}}, Y_{s_{2}}^{\textit{tgt}})$. This generation of paraphrase allows to fully utilize given annotation and effectively construct a dataset in target language with a limited amount of annotated data.

Similarly to image captioning task, we configured a well-designed prompt $P^{\textit{tgt}}$ for the annotation process, including an one-shot example. Specific prompts can be found in Appendix~\ref{sec:appendix-prompt-tst}. The process described in this section can be formulated as follows.

\begin{equation}
\label{eq1}
\begin{aligned}
    \{d_{1}^{\textit{tgt}},\;d_{2}^{\textit{tgt}}\}=\{(X_{s_{1}}^{\textit{tgt}}, Y_{s_{1}}^{\textit{tgt}}),\;(X_{s_{2}}^{\textit{tgt}}, Y_{s_{2}}^{\textit{tgt}})\} \\
    = \textrm{GPT}(P^{\textit{tgt}},\;(X^{\textit{eng}}, Y_{g}^{\textit{eng}}))
\end{aligned}
\end{equation}

We could acquire text style transfer dataset $D^{\textit{tgt}}$ in the target language through this process.

\section{Experiment}
\subsection{Experimental Design}
This section describes experimental design to validate the effectiveness of our GPT annotator in each tasks. We primarily assessed our method based on the performance of the model trained on the downstream task, which can serve as an indirect measure of the quality of synthesized dataset \cite{ye-etal-2022-zerogen}. Further implementation details can be found in Appendix~\ref{sec:appendix-impl-details}.

\subsubsection{Image Captioning Task}
To assess the cost-efficiency of our GPT annotator, we evaluated the proposed GPT annotator through three different image captioning datasets: Flickr8k \cite{rashtchian-etal-2010-collecting} dataset was constructed by annotating approximately 8,000 images collected from Flickr via MTurk. Flickr30k \cite{young-etal-2014-image} dataset is an extension of Flickr8k dataset, and it consisted of 30,000 images with captions acquired through crowdsourcing. MSCOCO \cite{lin2014microsoft, chen2015microsoft} dataset is an annotated dataset of more than 160,000 images.

As Flickr8k and Flickr30k datasets do not provide explicit validation and test sets, we divided them in the ratio of 8:1:1. For the MSCOCO dataset, we utilized the COCO 2014 split, which consists of approximately 82,000 training data, 40,000 validation data, and 40,000 test data. To validate the effectiveness of the proposed method, we set up a scenario with only one gold caption per image by selecting only one caption for the original dataset. 

BLEU \cite{papineni2002bleu}, ROUGE \cite{lin2004rouge}, and METEOR \cite{denkowski2014meteor} metrics were measured through the NLG-EVAL library \cite{sharma2017nlgeval} for evaluation. Additionally, we also employed BERTScore \cite{zhang2020bertscore} and BARTScore \cite{yuan2021bartscore} for model-based evaluation. For the MSCOCO dataset, the performance was evaluated through the official evaluation server.\footnote{\url{https://codalab.lisn.upsaclay.fr/competitions/7404}} For multilingual experiments, we adapted different datasets for each language, a subset of the aforementioned datasets with annotated captions. These datasets will be accordingly discussed in each section. We report the average performance of the model trained on three different random seeds, except the result on MSCOCO 2014 dataset.

\subsubsection{Text Style Transfer Task}

For text style transfer task, we conducted our experiments based on XFormal \cite{briakou-etal-2021-ola} dataset, which encompasses French, Brazilian Portuguese, and Italian. First, we selected 6,000 data for the GYAFC \cite{rao-tetreault-2018-dear} dataset, an English dataset that performs the same text formality style transfer, and translated them into each language using the NLLB \cite{costa2022no} model and Google Translator\footnote{\url{https://translate.google.com}} to build a baseline dataset. Second, we built a dataset with only 3,000 English data using our GPT Annotator as it generates two target language data for each English data. Using each dataset built by the Translation model and GPT Annotator respectively, we fine-tuned mBART \cite{tang2020multilingual} model to perform text style transfer task, and compared its performance and the formality of the generated text. Simliarly to image captioning task, NLG-EVAL library, as well as BERTScore and BARTScore were deployed for measuring metrics. Throughout the manuscript, we report the average performance of the model trained on three different random seeds.

\subsection{Cost-Efficiency of GPT Annotator}
\label{sec:exp-cost}

\begin{table}[h]
\resizebox{\columnwidth}{!}{%
\begin{tabular}{c|ccccc}
\Xhline{3\arrayrulewidth}
\textbf{Flickr8k}                                                           & BLEU  & ROUGE & METEOR & BERTS. & BARTS.  \\ \hline\hline
\begin{tabular}[c]{@{}c@{}}Human Annotator\\ w/ Limited Budget\end{tabular} & 28.96 & 38.76 & 17.83  & 0.7817 & -18.379 \\ \hline
Synonym Replacement                                                         & 30.30 & 38.61 & 17.61  & 0.7802 & -18.457 \\ \hline
Back-Translation                                                            & 30.02 & 39.02 & 17.32  & 0.7795 & -18.413 \\ \hline
HRQ-VAE                                                                     & 21.62 & 29.53 & 15.83  & 0.7542 & -18.641 \\ \hline
\begin{tabular}[c]{@{}c@{}}GPT Annotator\\ w/ GPT-3.5\end{tabular}          & 33.13 & 39.98 & 18.41  & 0.7892 & -18.374 \\ \Xhline{2\arrayrulewidth}
\textbf{Flickr30k}                                                          & BLEU  & ROUGE & METEOR & BERTS. & BARTS.  \\ \hline\hline
\begin{tabular}[c]{@{}c@{}}Human Annotator\\ w/ Limited Budget\end{tabular} & 25.72 & 34.14 & 15.66 & 0.7539 & -18.350  \\ \hline
Synonym Replacement                                                         & 26.78 & 35.28 & 15.54 & 0.7556 & -18.329  \\ \hline
Back-Translation                                                            & 27.32 & 36.70 & 15.67 & 0.7591 & -18.321 \\ \hline
HRQ-VAE                                                                     & 20.94 & 27.53 & 12.97  & 0.7385 & -18.542 \\ \hline
\begin{tabular}[c]{@{}c@{}}GPT Annotator\\ w/ GPT-3.5\end{tabular}          & 30.57 & 37.68 & 16.02 & 0.7669 & -18.298 \\ \Xhline{2\arrayrulewidth}
\textbf{MSCOCO 2014}                                                        & BLEU  & ROUGE & METEOR & BERTS. & BARTS.  \\ \hline\hline
\begin{tabular}[c]{@{}c@{}}Human Annotator\\ w/ Limited Budget\end{tabular} & 40.40 & 46.60 & 18.90  &  &  \\ \hline
Synonym Replacement                                                         & 45.10 & 50.30 & 23.90  &  &  \\ \hline
Back-Translation                                                            & 41.35 & 46.70 & 21.80  &  &  \\ \hline
HRQ-VAE                                                                     & 45.59 & 50.10 & 24.20  &  &  \\ \hline
\begin{tabular}[c]{@{}c@{}}GPT Annotator\\ w/ GPT-3.5\end{tabular}          & 46.38 & 50.40 & 24.50  &  &  \\ \Xhline{3\arrayrulewidth}
\end{tabular}
}
\caption{Experimental results to validate the cost-efficiency of the proposed GPT annotator. We only report BLEU, ROUGE, and METEOR for MSCOCO 2014 dataset as the official evaluation server does not provide BERTScore and BARTScore result.}
\label{tab-cost}
\end{table}

Based on the concept of a previous study \cite{wang-etal-2021-want-reduce}, we evaluated the difference in the performance of human annotators and GPT annotator under a fixed budget. The previous study \cite{rashtchian-etal-2010-collecting} suggested that it takes 0.05\$ to create five gold captions per image, which is equivalent to 0.01\$ for each gold caption. In the experiment, approximately 1000 tokens were used to generate annotated data pair.

According to this cost analysis, the method proposed in this study required 0.012\$ to generate one gold caption and four silver captions for each image using GPT-3.5, as it takes approximately 1,000 tokens to generate silver captions.\footnote{As of the time of this study, GPT-3.5 charged 0.002\$ per 1000 tokens. Currently, it charges 0.001\$ per 1000 tokens of prompt and 0.002\$ per 1000 tokens of generation.} Based on this configuration, it would cost approximately 76.8\$ to exploit GPT annotator to annotate the 6,400 images in the Flickr8k train set. In contrast, only 1,500 images can be annotated by purely human annotators under the same fixed budget. Similarly, for Flickr30k dataset, annotating 24,000 train data using the proposed method would cost approximately 288\$, whereas for the same amount, human annotators can only annotate 5,800 images to generate five gold captions. Following the same configuration, in the MSCOCO dataset, only 19,680 images can be annotated by human annotators under the budget that can annotate 82,000 images with GPT annotator.

Under this scenario, we compared the results of training the model by selecting only 1,500 fully human-annotated data from Flickr8k dataset, 5,800 fully human-annotated data from Flickr30k dataset, and 19,680 fully human-annotated data from MSCOCO dataset with the results obtained by training the model using the GPT-annotated data for the entire images of each dataset. Additionally, we also exploited other data augmentation baselines such as synonym replacement \cite{zhang2015character}, Back-Translation \cite{sennrich2016improving} and HRQ-VAE \cite{hosking2022hierarchical} to augment one gold data for extensive comparison.

Table~\ref{tab-cost} shows the results of the experiment. The experimental results suggest that under the same budget, annotating a larger number of images with one gold caption and multiple silver captions resulted in improved performance compared to annotating a smaller number of images with multiple gold captions using only human annotators. This outcome is consistent with the findings of previous work \cite{wang-etal-2021-want-reduce}, indicating the cost efficiency of GPT annotators, and indicates that these characteristics of GPT annotators are applicable to a wider range of tasks including image captioning. Furthermore, GPT annotator has shown superior performance against other augmentation baselines, suggesting that GPT annotator can generate better and diverse sentences.

\subsection{Multilingual Experiment}

\subsubsection{Korean Experiment}

Korean is a language that is attracting increasing attention owing to its approximately 80 million native speakers and rising Korean content. Nevertheless, the resource to fulfill this demand is limited \cite{gu-etal-2018-meta, sennrich-zhang-2019-revisiting, kim2021commonsense, sahoo-etal-2023-prejudice}. For example, there is no dedicated Korean dataset for the image captioning task. Although there are data that applied machine translation to existing English datasets, they are not fully open and have limited availability.\footnote{\url{https://aihub.or.kr} operated by the Korean government offers a machine-translated version of COCO captioning dataset; however, the public usage of this dataset is limited as it is only available to Korean citizens.}

Considering these characteristics of the Korean language, we first evaluated the multilingual ability of the proposed method based on Korean. In this experiment, we assessed the effectiveness of a Korean image captioning model which was trained on two separate datasets: the AiHub dataset, which applies machine translation to the English dataset, and the Korean dataset constructed by GPT-4 using the approach described in this study. Due to the absence of dedicated evaluation set for a fair comparison, human evaluation was conducted on 100 captions generated by each model from the test image set. The human evaluation was performed in accordance with the previously proposed protocol \cite{kasai-etal-2022-transparent}, and we report the average THUMB score of three Korean native speakers. 

Table~\ref{tab-kor-human} presents the results of the human evaluation. The outcomes of the evaluation indicate that the model trained on the dataset using GPT annotator performed better than the machine-translated dataset in terms of ratings by humans. In addition, our GPT annotator demonstrated a lower penalty on fluency, which suggests that our method generates more natural sentences.

\begin{table}[]
\resizebox{\columnwidth}{!}{%
\begin{tabular}{c|cccc}
\Xhline{3\arrayrulewidth}
\textbf{Korean}                                                      & Precision↑ & Recall↑ & Fluency↓ & THUMB↑ \\ \hline\hline
\begin{tabular}[c]{@{}c@{}}AiHub\\ (Machine-Translated)\end{tabular} & 4.3        & 4.09      & 0.03       & 4.17     \\ \hline
\begin{tabular}[c]{@{}c@{}}GPT Annotator\\ w/ GPT-4\end{tabular}     & 4.72       & 4.59      & 0.02       & 4.64     \\ \Xhline{3\arrayrulewidth}
\end{tabular}
}
\caption{Human evaluation results of the validation of the effectiveness of the proposed GPT annotator on Korean language. We follow the evaluation process and metric of THUMB \cite{kasai-etal-2022-transparent}, and report the average THUMB score of three Korean native speakers. Please refer to Appendix~\ref{sec:appendix-quant-kor} for quantitative analysis.}
\label{tab-kor-human}
\end{table}

These evaluation results confirmed that the model can achieve improved performance when trained with the dataset constructed using the method proposed in this study. Furthermore, as our GPT annotator generates five Korean captions using only one gold English caption by a human annotator, it is more cost-efficient compared to applying machine translation to five gold captions in English. Moreover, our GPT annotator has additional advantages that could ensure consistency in sentence structure compared to machine translation. Specifically, we instructed the annotator to generate sentences in the neutral form \begin{CJK}{UTF8}{mj}{(``-하다'')}\end{CJK} rather than the polite form \begin{CJK}{UTF8}{mj}{(``-합니다'')}\end{CJK} through the prompt. We can maintain consistency in tone and style of the dataset through this configuration, leading to better for the quality of the annotated data and reduce the need for post-processing and human intervention.

\subsubsection{Vietnamese Experiment}

Vietnamese also has more than 85 million native speakers, but suffering from lack of annotated data \cite{ngo-etal-2020-improving, huynh-etal-2022-vinli}. To demonstrate the versatility of our approach in another language, we expanded our experiments to Vietnamese. For the experiment, we adapted UiT-ViIC dataset \cite{lam2020uit}. This dataset consists of images selected from the MSCOCO dataset relating to sports, each with five Vietnamese captions manually annotated by a human annotator. We applied NLLB model and Google Translator to build a baseline by translating English captions from the original MSCOCO dataset into Vietnamese. Additionally, we adopted the data generated by HRQ-VAE in Section~\ref{sec:exp-cost} and translated them into Vietnamese using NLLB model.

\begin{table}[]
\resizebox{\columnwidth}{!}{%
\begin{tabular}{c|ccccc}
\Xhline{3\arrayrulewidth}
\textbf{Vietnamese}                                                  & BLEU  & ROUGE & METEOR & BERTS. & BARTS. \\ \hline\hline
\begin{tabular}[c]{@{}c@{}}Original\\ (Human-Annotated)\end{tabular} & 48.62 & 53.82 & 32.16 & 0.8309 & -14.511 \\ \hline
\begin{tabular}[c]{@{}c@{}}NLLB\\ (Machine-Translated)\end{tabular}  & 31.76 & 40.49 & 26.61 & 0.8114 & -14.645  \\ \hline
\begin{tabular}[c]{@{}c@{}}HRQ-VAE + NLLB\end{tabular}               & 21.26 & 28.64 & 23.48 & 0.7720 & -15.342  \\ \hline
Google Translator                                                    & 37.22 & 46.24 & 26.86 & 0.8196 & -14.534  \\ \hline
\begin{tabular}[c]{@{}c@{}}GPT Annotator \\ w/ GPT-4\end{tabular}    & 41.32 & 47.83 & 30.57 & 0.8235 & -14.537 \\
\Xhline{3\arrayrulewidth}
\end{tabular}
}
\caption{Experimental results in Vietnamese based on UiT-ViIC dataset.}
\label{tab-vie}
\end{table}

Table~\ref{tab-vie} presents the results on Vietnamese. The experimental result suggests that our approach is valid in Vietnamese, leading to better performance of the model compared to a machine translation-based approach.

\subsubsection{Polish Experiment}

Polish is another language that has challenge of low-resource language \cite{dadas-etal-2020-evaluation, augustyniak2022way}. To further validate our method's applicability, we also conducted experiments on the AIDe dataset for Polish \cite{wroblewska-2018-polish}. This dataset is composed of 1,000 images selected from the Flickr8k dataset, each with two human-annotated captions in Polish. For this experiment, we configured our prompt to generate two caption pairs for each image. Similarly to Vietnamese experiment, for the Polish translation baseline, we utilized the NLLB model and Google Translator to translate two English captions from the original Flickr8k dataset into Polish. We also adopted the data generated by HRQ-VAE in Section~\ref{sec:exp-cost} and translated them into Polish using NLLB model.

\begin{table}[]
\resizebox{\columnwidth}{!}{%
\begin{tabular}{c|ccccc}
\Xhline{3\arrayrulewidth}
\textbf{Polish}                                                      & BLEU  & ROUGE & METEOR & BERTS. & BARTS. \\ \hline\hline
\begin{tabular}[c]{@{}c@{}}Original\\ (Human-Annotated)\end{tabular} & 8.68 & 19.38 & 9.38 & 0.7405 & -18.162 \\ \hline
\begin{tabular}[c]{@{}c@{}}NLLB\\ (Machine-Translated)\end{tabular}  & 4.14 & 14.46 & 6.78 & 0.6466 & -18.279  \\ \hline
\begin{tabular}[c]{@{}c@{}}HRQ-VAE + NLLB\end{tabular}               & 3.21 & 13.15 & 5.99 & 0.6495 & -18.331 \\ \hline
Google Translator                                                    & 4.64 & 14.14 & 6.91 & 0.6507 & -18.244 \\ \hline
\begin{tabular}[c]{@{}c@{}}GPT Annotator \\ w/ GPT-4\end{tabular}    & 5.17 & 18.90 & 8.92 & 0.6962 & -18.197  \\ 
\Xhline{3\arrayrulewidth}
\end{tabular}
}
\caption{Experimental results in Polish based on AIDe dataset.}
\label{tab-pl}
\end{table}

Table~\ref{tab-pl} indicates the results on Polish. The experimental result demonstrates the effectiveness of our approach, showcasing not just better performance compared to translation baseline but also comparable performance to human-annotated data.

\subsection{Text Style Transfer Experiment}

\begin{table}[]
\resizebox{\columnwidth}{!}{%
\begin{tabular}{c|cccccc}
\Xhline{3\arrayrulewidth}
\textbf{French}                                                     & BLEU  & ROUGE & METEOR & BERTS. & BARTS. & Formality      \\ \hline\hline
\begin{tabular}[c]{@{}c@{}}NLLB\\ (Machine-Translated)\end{tabular} & 48.59 & 50.26 & 31.42 & 0.8103 & -17.596 &  72.37 \\ \hline
Google Translator                                                   & 51.69 & 54.02 & 32.62 & 0.8076 & -17.541      & 75.38 \\ \hline
\begin{tabular}[c]{@{}c@{}}GPT Annotator \\ w/ GPT-4\end{tabular}   & 54.81 & 56.83 & 33.98  & 0.8175 & -17.519 & \textbf{85.12} \\ \Xhline{2\arrayrulewidth}
\textbf{Brazilian Portuguese}                                       & BLEU  & ROUGE & METEOR & BERTS. & BARTS. & Formality      \\ \hline\hline
\begin{tabular}[c]{@{}c@{}}NLLB\\ (Machine-Translated)\end{tabular} & 52.73 & 55.81 & 32.44  & 0.8286 & -18.955 & 68.58 \\ \hline
Google Translator                                                   & 55.98 & 57.74 & 34.19  & 0.8318 & -18.938 & 74.27    \\ \hline
\begin{tabular}[c]{@{}c@{}}GPT Annotator \\ w/ GPT-4\end{tabular}   & 57.94 & 60.72 & 35.60  & 0.8363 & -18.864 & \textbf{79.21} \\ \Xhline{2\arrayrulewidth}
\textbf{Italian}                                                    & BLEU  & ROUGE & METEOR & BERTS. & BARTS. & Formality      \\ \hline\hline
\begin{tabular}[c]{@{}c@{}}NLLB\\ (Machine-Translated)\end{tabular} & 47.97 & 49.34 & 30.12  & 0.7839 & -18.843 &  68.03 \\ \hline
Google Translator                                                   & 49.13 & 51.73 & 30.89  & 0.7873  & -18.805   & 71.86 \\ \hline
\begin{tabular}[c]{@{}c@{}}GPT Annotator \\ w/ GPT-4\end{tabular}   & 52.34 & 53.71 & 32.02 & 0.7994 & -18.702 & \textbf{74.29} \\
\Xhline{3\arrayrulewidth}
\end{tabular}
}
\caption{Experimental results on text style transfer in French, Brazilian Portuguese, and Italian.}
\label{tab-tst}
\end{table}

Table~\ref{tab-tst} presents the experimental result of our GPT annotator for text style transfer task in French, Brazilian Portuguese, and Italian. The results not only highlight the performance of our GPT Annotator with fewer original human-annotated samples but also underscore its ability to enhance text formality against translation. This achievement was possible through the consistent generation of sentences with formal and informal styles, owing to the flexibility of LLMs and instructible prompts.

\subsection{Employing GPT Annotator for Dataset Construction}

Latvian, Estonian, and Finnish have approximately 1.5, 1.1, and 4.8 million native speakers, which make them hard to hire annotators and construct datasets. To address the practical challenges in the field of data annotation, we constructed an image captioning dataset in these languages, which did not have any image captioning dataset, using our GPT annotator. We first selected 3,850 images and their English captions from the MSCOCO dataset and splited them into 2,695 train images, 924 validation images, and 231 test images, following the configuration of the Vietnamese UiT-ViIC dataset. 

To build a baseline, we utilized NLLB and Google Translator to translate the English caption of each training image, similar to previous experiments. The validation and test captions were constructed by translating using mBART model, for a fair comparison.

\begin{table}[]
\resizebox{\columnwidth}{!}{%
\begin{tabular}{c|ccccc}
\Xhline{3\arrayrulewidth}
\textbf{Latvian}                                                     & BLEU  & ROUGE & METEOR & BERTS. & BARTS. \\ \hline\hline 
\begin{tabular}[c]{@{}c@{}}NLLB\\ (Machine-Translated)\end{tabular}  & 6.39  & 17.53 & 10.13  & 0.6803 & -16.061   \\ \hline
\begin{tabular}[c]{@{}c@{}}HRQ-VAE + NLLB\end{tabular}               & 5.14  & 16.61 & 10.21  & 0.6728 & -16.127  \\ \hline
Google Translator                                                    & 8.53  & 17.09 & 10.67  & 0.6848 & -16.067  \\ \hline
\begin{tabular}[c]{@{}c@{}}GPT Annotator \\ w/ GPT-4\end{tabular}    & 10.35 & 18.61 & 10.79  & 0.6911  & -16.054   \\ \Xhline{2\arrayrulewidth}
\textbf{Estonian}                                                    & BLEU  & ROUGE & METEOR & BERTS. & BARTS. \\ \hline\hline 
\begin{tabular}[c]{@{}c@{}}NLLB\\ (Machine-Translated)\end{tabular}  & 4.97 & 13.12 & 7.89 & 0.6893 & -15.409  \\ \hline
\begin{tabular}[c]{@{}c@{}}HRQ-VAE + NLLB\end{tabular}               & 3.37 & 7.84  & 5.87 & 0.6876 & -15.409  \\ \hline
Google Translator                                                    & 6.04 & 12.51 & 8.75 & 0.7008 & -15.408  \\ \hline
\begin{tabular}[c]{@{}c@{}}GPT Annotator \\ w/ GPT-4\end{tabular}    & 6.62 & 13.47 & 9.22 & 0.7050 & -15.407  \\ \Xhline{2\arrayrulewidth}
\textbf{Finnish}                                                     & BLEU  & ROUGE & METEOR & BERTS. & BARTS. \\ \hline\hline 
\begin{tabular}[c]{@{}c@{}}NLLB\\ (Machine-Translated)\end{tabular}  & 4.19 & 10.43 & 7.74 & 0.7122 & -16.392  \\ \hline
\begin{tabular}[c]{@{}c@{}}HRQ-VAE + NLLB\end{tabular}               & 3.74 & 10.23 & 7.06 & 0.6965 & -16.401  \\ \hline
Google Translator                                                    & 4.28 & 10.84 & 7.88 & 0.7128 & -16.394  \\ \hline
\begin{tabular}[c]{@{}c@{}}GPT Annotator \\ w/ GPT-4\end{tabular}    & 4.96 & 12.29 & 8.64 & 0.7143 & -16.389  \\ \Xhline{2\arrayrulewidth}
\end{tabular}
}
\caption{Experimental results of our constructed dataset in Latvian.}
\label{tab-lt}
\end{table}

Table~\ref{tab-lt} clearly showcases the efficiency of our GPT annotator when human-annotated data is scarce, as observed in case of these low-resource languages. The human investigation of annotated data remains for future work. We plan to release the training, validation, and testing datasets for wider access and further study. This experimental result demonstrates the possibility of the GPT annotator to easily construct dataset in any designated language, enhancing the accessibility of various languages.

\section{Conclusion}
In this study, we have demonstrated the possibility of exploiting LLM as a multilingual assistant annotator by generating multiple silver data from a single gold data in different languages. The experimental results showcased that the proposed method is cost-efficient compared to entirely human annotation, and can be effectively employed to construct datasets in various languages and tasks.

The approach described in this work can be widely adapted to various languages, as it utilizes the multilingual fluency and flexibility of LLMs. We constructed an image captioning in Latvian as a practical application of our GPT annotator. Furthermore, the cost-efficiency of the GPT annotator suggested in this paper will be improved in the future, as the price of LLMs is expected to decline as recent cost reductions of GPT-3.5 and GPT-4 have shown. Future studies will focus on improving the proposed method by utilizing the image inception ability and expanding this method to other tasks.

\section*{Limitations}
Extreme low-resource languages may still encounter difficulty producing high-quality sentences even with the use of GPT-4. To examine the responses of GPT-4 in translating into extremely low-resource languages, we conducted an error analysis in two extremely low-resource languages, Basque and Māori. Basque has a small amount of speakers, and it is also a unique language isolate, that does not have a distinct relationship with other languages such as Spanish and French, making it harder to process. Māori has a very small amount of language users, posing a challenge as an extremely low-resource language. Please refer to Appendix~\ref{sec:appendix-error} for the analysis result.

Additionally, the approach demonstrated in this work generates silver sentences as paraphrases of the given gold sentences, thus they might not fully capture the information that exists in the image but is not mentioned in the gold sentences. Consequently, the gold captions produced by multiple human annotators can be more diverse than silver captions. To address this issue, human annotators could create gold captions that contain as much detailed and diverse information as possible while constructing a new dataset through this method. 

\section*{Ethics Statement}
As this work proposes the utilization of LLMs as an assistant data annotator and for the automatic generation of sentences, it may suffer from the potential bias of LLMs. To mitigate this concern, we added explicit instructions to prevent the generation of biased sentences in the prompts. However, the human supervisor is still essential to examine and validate the absence of biased expressions in the generated data. Specifically, the human supervisor should ensure that there is not any biased gold sentence produced by the human annotator, as it directly affects the bias of generated sentences using LLMs.

Furthermore, in addition to the error analysis presented in the previous section, we have conducted supplementary error analysis on Basque and Māori languages in Appendix~\ref{sec:appendix-error-ethics}. This additional investigation aims to explore the potential ethical biases exhibited by GPT-4. Our findings suggest that GPT-4 may exhibit unexpected ethical biases, particularly in extremely low-resource languages, where its knowledge about the language may be limited compared to high-resource languages such as English.

\section*{Acknowledgements}
This research was supported by Basic Science Research Program through the National Research Foundation of Korea(NRF) funded by the Ministry of Education(NRF-2022R1C1C1008534), and Institute for Information \& communications Technology Planning \& Evaluation (IITP) through the Korea government (MSIT) under Grant No. 2021-0-01341 (Artificial Intelligence Graduate School Program, Chung-Ang University).

\bibliography{custom}

\appendix
\section{Implementation Details}
\label{sec:appendix-impl-details}
\subsection{Model Implementation}
PyTorch \cite{paszke2019pytorch} and Huggingface Transformers library \cite{wolf-etal-2020-transformers} have been employed for the implementation process. 

For image captioning task, Vision Transformer (ViT) \cite{dosovitskiy2021image} and Transformer \cite{vaswani2017attention} were deployed as the encoder and decoder of the model, respectively. Particularly, pretrained \textit{vit\_b\_16} from torchvision library \cite{torchvision2016} was adapted as an encoder, and the decoder consisted of 12 heads and 12 layers, with a hidden layer size and embedding layer size of 768.

For text style transfer task, we fine-tuned \textit{mbart-50-large} model using each dataset to convert informal text into formal text. Additionally, we separately trained another mBART model as formality classifier using XFormal training data for each language to measure the formality of the generated text. The text formality was measured by the average logit of the classifier.

Every model was trained using AdamW \cite{loshchilov2018decoupled} with a batch size of 16 and a learning rate of 5e-5 through 10 epochs, while the weight decay of the optimizer was set to 1e-5, and a CosineAnnealingLR \cite{loshchilov2017sgdr} scheduler was deployed.

\subsection{GPT Annotator Implementation}
We utilized the official API from OpenAI to implement the proposed GPT annotator. The versions of the models used are \textit{gpt-3.5-turbo-0301} and \textit{gpt-4-0314}, respectively. The prompts used can be found in Appendix~\ref{sec:appendix-prompt}. If an error occurred while generating an annotation using a given prompt, the API was called again with a patience of three times. If this patience was exceeded, the data pair was excluded from the annotation process.

\subsection{Further Details}
We employed the \textit{facebook/nllb-200-distilled-600M} model, which comprises 600M parameters, to create a training dataset using the NLLB baseline. Similarly, we utilized the \textit{facebook/mbart-large-50-many-to-many-mmt} model, with approximately 611M parameters, to construct validation and test sets for Latvian, Estonian, and Finnish. This choice was made to ensure a fair and equitable comparison between the baseline models and our proposed GPT annotator. For evaluation with BERTScore \cite{zhang2020bertscore} and BARTScore \cite{yuan2021bartscore}, we exploited \textit{bert-base-multilingual-cased} and \textit{facebook/mbart-large-50}, respectively. Note that we reported BERTScore-F1 in the manuscript.

Label smoothing \cite{szegedy2016rethinking} was applied with a smoothing epsilon of 0.05. The training procedure was conducted on a single Nvidia RTX 3090 GPU.

For the tokenizing of text input, we employed tokenizer of pre-trained model available on Huggingface for each language. Specifically, \textit{facebook/bart-base}, \textit{cosmoquester/bart-ko-base}, \textit{vinai/bartpho-syllable}, \textit{sdadas/polish-bart-base}, and \textit{joelito/legal-latvian-roberta-base}, \textit{tartuNLP/EstBERT}, \textit{TurkuNLP/bert-base-finnish-uncased-v1} were adapted as the tokenizer for English, Korean, Vietnamese, Polish, Latvian, Estonian, and Finnish. For text style transfer task, as it is based on \textit{facebook/mbart-large-50} model, each language shares same tokenizer.

For the test procedure of the Flickr8k and Flickr30k datasets, all five available human-annotated captions of the test set were utilized as reference sentences for evaluation. Beam search \cite{freitag-al-onaizan-2017-beam} was applied as a decoding strategy to generate sentences at inference time, with a beam size of 5.

\begin{table*}[h]
\centering
\scalebox{0.9}{%
\begin{tabular}{c|ccc|ccc}
\Xhline{3\arrayrulewidth}
Evaluation Method                                                    & \multicolumn{3}{c|}{Validation Set (Korean)}                                       & \multicolumn{3}{c}{Test Set (Translated to English)}                              \\
Metric                                                               & \multicolumn{1}{l}{BLEU} & \multicolumn{1}{l}{ROUGE} & \multicolumn{1}{l|}{METEOR} & \multicolumn{1}{l}{BLEU} & \multicolumn{1}{l}{ROUGE} & \multicolumn{1}{l}{METEOR} \\ \Xhline{2\arrayrulewidth}
\begin{tabular}[c]{@{}c@{}}AiHub\\ (Machine-Translated)\end{tabular} & 11.20                    & 20.64                     & 19.41                       & 34.85                    & 41.60                     & 19.80                      \\ \hline
\begin{tabular}[c]{@{}c@{}}GPT Annotator\\ w/ GPT-4\end{tabular}     & 7.01                        & 15.84                         & 18.56                           & 32.70                        & 39.90                         & 19.20                          \\ \Xhline{3\arrayrulewidth}
\end{tabular}
}
\caption{Quantitative experimental results of the machine-translated dataset and proposed GPT annotator on Korean language. The left column (`Validation Set') refers to the inference result of the validation set provided in Korean. The right column (`Test Set') is the evaluation result of the Korean model, but as there is no Korean data for the test set, we translated the Korean inference result into English and uploaded it to the official evaluation server.}
\label{tab-kor}
\end{table*}

\begin{table}[h!]
\resizebox{\columnwidth}{!}{
\begin{tabular}{ccccc}
\Xhline{3\arrayrulewidth}
\multicolumn{1}{c|}{Metric}    & Precision & Recall & Fluency & THUMB \\ \Xhline{2\arrayrulewidth}
\multicolumn{1}{c|}{Human \#1} & 4.61      & 4.26   & 0.01    & 4.43  \\ \hline 
\multicolumn{1}{c|}{Human \#2} & 4.3       & 4.21   & 0.05    & 4.21  \\ \hline 
\multicolumn{1}{c|}{Human \#3} & 4.62      & 4.56   & 0.01    & 4.58  \\ \Xhline{3\arrayrulewidth} 
\end{tabular}}
\caption{For transparency of human evaluation, we report the average value of each metric as rated by three raters.}
\label{tab-kor-human-2}
\end{table}

\section{GPT Annotator Software}

In order to streamline the annotation process outlined in this paper, we have developed specialized software tailored for multilingual data annotation, leveraging OpenAI GPT models. This software currently supports tasks such as image captioning, text style transfer, and machine translation. Although these functionalities are not discussed in detail in this paper due to space constraints, they are available within the software.

The annotator software takes a JSON file as input and generates a new JSON file containing multilingual annotations in the target language. This is achieved by utilizing the specified prompt and the chosen version of the GPT model. Moreover, the software is designed to facilitate faster data annotation through multiprocessing capabilities. For a more comprehensive understanding of the software's functionality, please refer to the attached code. 

\section{Quantitative Experiments on Korean}
\label{sec:appendix-quant-kor}

We have included the human evaluation results in Table~\ref{tab-kor-human} within the main manuscript. This was done because there is no dedicated evaluation set available in Korean, which is essential for a fair evaluation. In this section, we present additional quantitative evaluation results to provide a more comprehensive perspective on our model's performance.

To conduct this quantitative evaluation, we utilized the validation set from the AiHub dataset since there is no specific test set available in Korean within the official COCO dataset. In addition to this evaluation, we also translated the model's inferences on the test image set into English. This allowed us to assess the model's performance on the test set using the official evaluation server. The quantitative analysis results are presented in Table~\ref{tab-kor}.

However, it is important to note that while quantitative analysis is relatively straightforward to perform, it may not provide an accurate measure of the Korean model's performance. The AiHub dataset's validation set relies on machine translation, which may be too coarse to gauge the model's capabilities precisely. Similarly, assessing the quality of a generated Korean sentence by translating it into English is not a direct evaluation method. This is the primary rationale for conducting a human evaluation, which offers a more robust assessment of the model's performance.

\section{Detailed Information on Human Evaluation}

\begin{figure*}[h]
    \centering
    \includegraphics[width=1\textwidth]{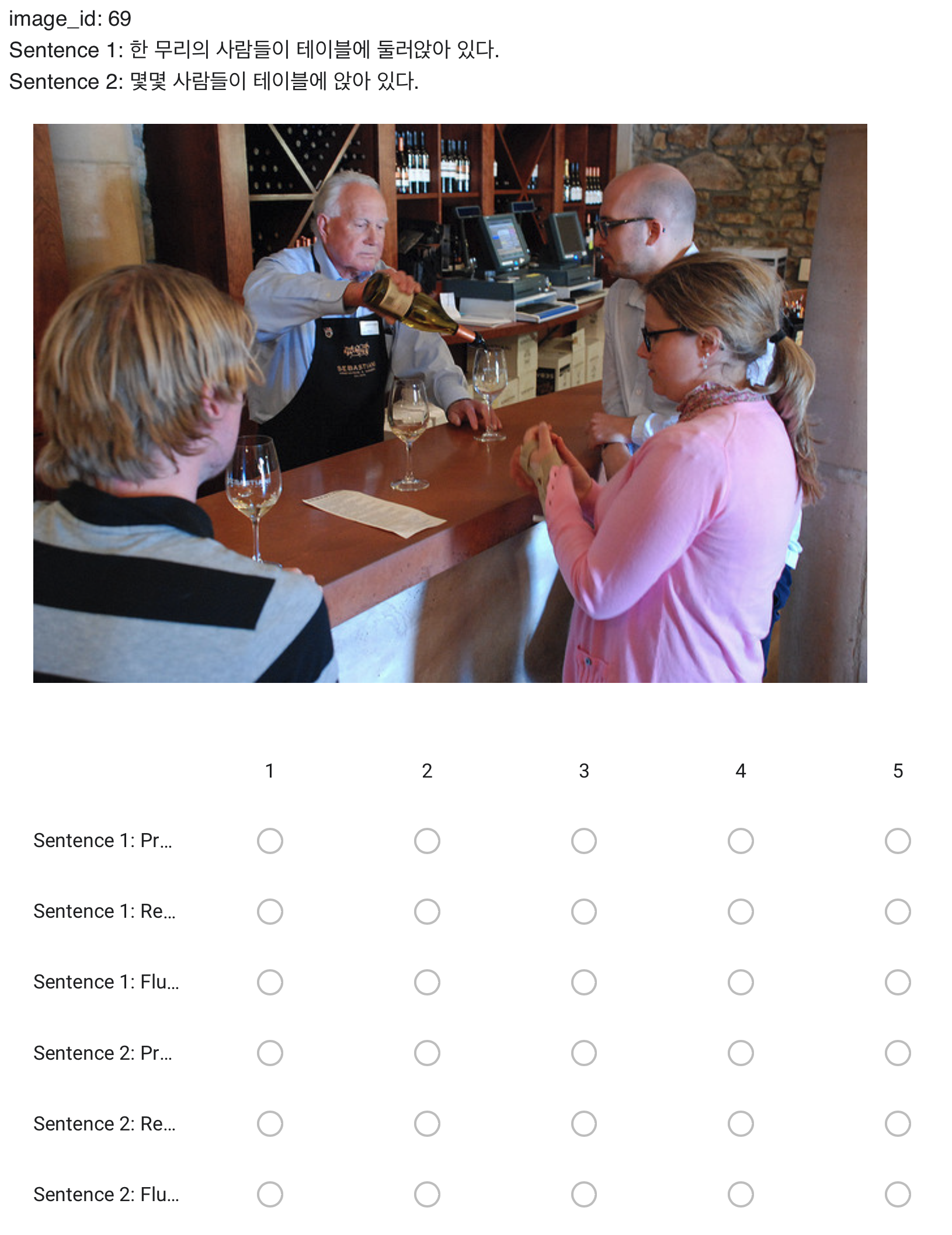}
    \caption{The screenshot of human evaluation form. Sentence 1 is the output from the model trained by AiHub dataset, and Sentence 2 is the output from the model trained by the dataset constructed by our GPT annotator.}
\label{fig-humaneval}
\end{figure*}

Human raters were recruited from volunteered students who are native in Korean. Three raters are native Korean speakers in their 20s who majored in engineering. The detailed information about THUMB score \cite{kasai-etal-2022-transparent}, the metric used in this study for the assessment of the generated caption, was provided to raters. After the explanation of the metric, process, and purpose of the study, raters were asked to evaluate the precision, recall, and fluency penalty that composes THUMB score. Figure~\ref{fig-humaneval} is a screenshot as an example of the human evaluation form. To prevent rater fatigue, We instructed them to pause the evaluation process if they felt exhausted and not to finish it all at once. 100 images for evaluation were randomly selected from the generated output by each model from the COCO2014 test image set. Table~\ref{tab-kor-human-2} shows the average evaluation result of each rater.

\newpage
\onecolumn
\section{Case Analysis}

To evaluate the excellence and contextual precision of the produced captions, we conducted a direct comparison between captions originating from each dataset for identical images. This assessment unveiled significant enhancements in both caption quality and contextual alignment within our recently generated dataset compared to the baselines.

\subsection{Korean Analysis}

\begin{itemize} 
    \item \textbf{Quality of Generated Sentence}
    \begin{itemize}
        \item MSCOCO Image ID: \texttt{237944}
        \begin{itemize}
            \item English Reference: \\ A person under a dryer wearing a towel
            \item AiHub (Machine-Translated): \\ \begin{CJK}{UTF8}{mj}{드레이더}\end{CJK} (\textit{Drader} - This word does not exist in Korean.)
            \item GPT Annotator w/ GPT-4: \\ \begin{CJK}{UTF8}{mj}{수건을 두른 사람이 드라이어 아래에 있다.}\end{CJK} (\textit{A person with a towel is under the dryer.})
        \end{itemize}
    \end{itemize}
    \begin{itemize}
        \item MSCOCO Image ID: \texttt{215878}
        \begin{itemize}
            \item English Reference: \\ A white microwave oven a pot holder and some books
            \item AiHub (Machine-Translated): \\ \begin{CJK}{UTF8}{mj}{하얀 전자 레인지에 냄비 뚜껑과 책 몇권을 넣어}\end{CJK} (\textit{Put a pot lid and some books in a white microwave})
            \item GPT Annotator w/ GPT-4: \\ \begin{CJK}{UTF8}{mj}{하얀 전자레인지 오븐, 냄비 받침이랑 몇 권의 책들이 있다.}\end{CJK} (\textit{There is a white microwave oven, pot holders, and some books.})
        \end{itemize}
    \end{itemize}
    \item \textbf{Context of Generated Sentence}
    \begin{itemize}
        \item MSCOCO Image ID: \texttt{190556}
        \begin{itemize}
            \item English Reference: \\ Close up images of bikes parked next to the highway.
            \item AiHub (Machine-Translated): \\ \begin{CJK}{UTF8}{mj}{고속 도로 옆에 주차된 자전거의 이미지를 닫아라.}\end{CJK} (\textit{Close the image of a bicycle parked on the side of the high way.})
            \item GPT Annotator w/ GPT-4: \\ \begin{CJK}{UTF8}{mj}{고속도로 옆에 주차된 자전거의 근접한 이미지들이다.}\end{CJK} (\textit{Close-up images of a bicycle parked on the side of the highway.})
        \end{itemize}
    \end{itemize}
    \begin{itemize}
        \item MSCOCO Image ID: \texttt{273929}
        \begin{itemize}
            \item English Reference: \\ A far away shot of Big Ben and the nearby complex.
            \item AiHub (Machine-Translated): \\ \begin{CJK}{UTF8}{mj}{멀리서 빅 벤과 인근 콤플렉스를 총으로 쐈어요}\end{CJK} (\textit{I shot Big Ben and the nearby complex from a distance with a gun})
            \item GPT Annotator w/ GPT-4: \\ \begin{CJK}{UTF8}{mj}{빅 벤과 인근 건물들을 멀리서 찍은 사진이다.}\end{CJK} (\textit{This is a photo of Big Ben and nearby buildings from a distance.})
        \end{itemize}
    \end{itemize}
\end{itemize}

\subsection{Vietnamese Analysis}

\begin{itemize} 
    \item \textbf{Quality of Generated Sentence}
    \begin{itemize}
        \item MSCOCO Image ID: \texttt{213669}
        \begin{itemize}
            \item English Reference: \\ A young man holding a tennis racquet on a tennis court.
            \item Vietnamese Reference: \\ \foreignlanguage{vietnamese}{Người đàn ông đang cầm vợt tennis chạy tới đánh bóng.} (\textit{A man holding a tennis racket runs to hit the ball.})
            \item NLLB (Machine-Translated): \\ \foreignlanguage{vietnamese}{một người đàn ông đứng trên một thức ăn với một tên lửa} (\textit{a man standing on a food with a rocket})
            \item GPT Annotator w/ GPT-4: \\ \foreignlanguage{vietnamese}{Một người trẻ tuổi đang ở trên sân tennis với cây vợt trong tay.} (\textit{A young person is on the tennis court with a racket in his hand.})
        \end{itemize}
    \end{itemize}
\end{itemize}

\subsection{Polish Analysis}

\begin{itemize} 
    \item \textbf{Context of Generated Sentence}
    \begin{itemize}
        \item Flickr File Name: \\ \texttt{1153704539\_542f7aa3a5}
        \begin{itemize}
            \item English Reference: \\ A girl playing trumpet in a marching band.
            \item Polish Reference: \\ Dziewczyna w sportowym stroju i czapce z daszkiem stoi na trawniku i gra na trąbce w towarzystwie innych muzyków. (\textit{A girl in sports clothes and a baseball cap stands on the lawn and plays the trumpet in the company of other musicians.})
            \item NLLB (Machine-Translated): \\ Dziewczyna grająca na trąbce w zespole. (\textit{A girl playing the trumpet in a band.})
            \item GPT Annotator w/ GPT-4: \\ Dziewczyna grająca na trąbce w orkiestrze marszowej. (\textit{A girl playing the trumpet in the march orchestra.})
        \end{itemize}
    \end{itemize}
    \item \textbf{Quality of Generated Sentence}
    \begin{itemize}
        \item Flickr File Name: \\ \texttt{1386251841\_5f384a0fea}
        \begin{itemize}
            \item English Reference: \\ A woman is looking at dressed, headless mannequins in a store display.
            \item Polish Reference: \\ Kobieta ogląda wystawę z ubranymi w damskie stroje manekinami. (\textit{A woman looks at an exhibition with mannequins dressed in women's clothes.})
            \item NLLB (Machine-Translated): \\ Kobieta patrzy na ubrane, bezgłowe manieki w sklepach. (\textit{A woman looks at clothed, headless maniacs in stores.})
            \item GPT Annotator w/ GPT-4: \\ Kobieta patrzy na ubrane, bezgłowe manekiny w wystawie sklepowej. (\textit{A woman looks at clothed, headless mannequins in a store window.})
        \end{itemize}
    \end{itemize}
    \begin{itemize}
        \item Flickr File Name: \\ \texttt{1387785218\_cee67735f5}
        \begin{itemize}
            \item English Reference: \\ A child pushes a doll in a baby carriage.
            \item Polish Reference: \\ Dziecko idzie drogą, prowadząc przed sobą mały wózek z lalką. (\textit{A child walks along the road, pushing a small stroller with a doll in front of him.})
            \item NLLB (Machine-Translated): \\ Mała dziewczynka wpychająca koło lalki (\textit{A little girl pushing a doll wheel})
            \item GPT Annotator w/ GPT-4: \\ Dziecko popycha lalkę w wózku dla dzieci. (\textit{A child pushes a doll in a baby stroller.})
        \end{itemize}
    \end{itemize}
    \begin{itemize}
        \item Flickr File Name: \\ \texttt{1499495021\_d295ce577c}
        \begin{itemize}
            \item English Reference: \\ A dark haired woman wearing a brown jacket and fatigue bottoms and a balding man wearing a green sweater and blue jeans with a fishing pole, stand at the foot of the surf.
            \item Polish Reference: \\ Mężczyzna stoi z wędką nad brzegiem wody, a obok niego stoi kobieta. (\textit{A man stands with a fishing rod at the water's edge and a woman stands next to him.})
            \item NLLB (Machine-Translated): \\ Czarnowłosa kobieta w brązowej kurtce i zmęczonym dnie i łysy mężczyzna w zielonej swetrze i niebieskich dżinsów z palcem rybaczem, stoją u stóp pływu. (\textit{A black-haired woman in a brown jacket and a tired bottom and a bald man in a green sweater and blue jeans with a fisherman's finger, stand at the foot of the tide.})
            \item GPT Annotator w/ GPT-4: \\ Kobieta o ciemnych włosach, ubrana w brązową kurtkę i spodnie w kamuflaż, oraz łysiejący mężczyzna w zielonym swetrze i niebieskich dżinsach z wędką, stoją u podnóża fali. (\textit{A woman with dark hair, wearing a brown jacket and camouflage pants, and a balding man in a green sweater and blue jeans with a fishing rod, stand at the foot of a wave.})
        \end{itemize}
    \end{itemize}
    \begin{itemize}
        \item Flickr File Name: \\ \texttt{146098876\_0d99d7fb98}
        \begin{itemize}
            \item English Reference: \\ A boy and three girls in blue school uniforms walk down a dirt-covered road.
            \item Polish Reference: \\ Chłopiec i trzy dziewczynki w mundurkach idą, niosąc zeszyty. (\textit{A boy and three girls in uniforms are walking, carrying notebooks.})
            \item NLLB (Machine-Translated): \\ Chłopak i trzy dziewczyny w niebieskich mundurach szli po błędnej drodze. (\textit{A boy and three girls in blue uniforms were walking on the wrong path.})
            \item GPT Annotator w/ GPT-4: \\ Chłopiec i trzy dziewczyny w niebieskich mundurkach szkolnych idą po drodze pokrytej brudem. (\textit{A boy and three girls in blue school uniforms are walking on a road covered with dirt.})
        \end{itemize}
    \end{itemize}
\end{itemize}

\subsection{Latvian Analysis}

\begin{itemize} 
    \item \textbf{Quality of Generated Sentence}
    \begin{itemize}
        \item MSCOCO Image ID: \texttt{46544}
        \begin{itemize}
            \item English Reference: \\ A woman playing tennis on a tennis court.
            \item NLLB (Machine-Translated): \\ Sieva tenisā tenisā. (\textit{Tennis wife in tennis.})
            \item GPT Annotator w/ GPT-4: \\ Sieviete spēlē tenisu tenisa kortā. (\textit{A woman plays tennis on a tennis court.})
        \end{itemize}
    \end{itemize}
    \begin{itemize}
        \item MSCOCO Image ID: \texttt{43960}
        \begin{itemize}
            \item English Reference: \\ A boy catching a ball while another boy holds a bat.
            \item NLLB (Machine-Translated): \\ Puikas, kas ieņem lopu, kamēr cits puikas, kas drīkst pieņemt lopu. (\textit{Boys who take livestock, while other boys who are allowed to accept livestock.})
            \item GPT Annotator w/ GPT-4: \\ Zēns noķer balls, kamēr cits zēns tur nūju. (\textit{A boy catches the ball while another boy holds the stick.})
        \end{itemize}
    \end{itemize}
    \begin{itemize}
        \item MSCOCO Image ID: \texttt{47813}
        \begin{itemize}
            \item English Reference: \\ There are four people playing tennis in doubles.
            \item NLLB (Machine-Translated): \\ Divās grupās spēlē četri cilvēki. (\textit{Four people play in two groups.})
            \item GPT Annotator w/ GPT-4: \\ Četri cilvēki spēlē tenisu dubultspēlēs. (\textit{Four people play tennis in doubles.})
        \end{itemize}
    \end{itemize}
\end{itemize}

\subsection{Estonian Analysis}

\begin{itemize} 
    \item \textbf{Quality of Generated Sentence}
    \begin{itemize}
        \item MSCOCO Image ID: \texttt{1596} 
        \begin{itemize}
            \item English Reference: \\ A person swing a tennis racket at a tennis ball.
            \item NLLB (Machine-Translated): \\ Üks inimene käigub tennisepalli peal tennise racket. (\textit{One person moves a tennis racket on top of a tennis ball.})
            \item GPT Annotator w/ GPT-4: \\ Inimene lööb tennise reketiga tennisepalli. (\textit{A person hits a tennis ball with a tennis racket.})
        \end{itemize}
    \end{itemize}
    \begin{itemize}
        \item MSCOCO Image ID: \texttt{35818}
        \begin{itemize}
            \item English Reference: \\ A group of boys play soccer in a grassy field.
            \item NLLB (Machine-Translated): \\ Grupp poisid mängib jalgpalli mägedes. (\textit{A group of boys plays football in the mountains.})
            \item GPT Annotator w/ GPT-4: \\ Poiste grupp mängib jalgpalli rohusel väljakul. (\textit{A group of boys plays football on a green field.})
        \end{itemize}
    \end{itemize}
    \begin{itemize}
        \item MSCOCO Image ID: \texttt{65500}
        \begin{itemize}
            \item English Reference: \\ Two sets of people are at a tennis net.
            \item NLLB (Machine-Translated): \\ Kaks inimest on tennistöö juures. (\textit{Two people are at tennis work.})
            \item GPT Annotator w/ GPT-4: \\ Kaks inimeste rühma on tennisevõrgu juures. (\textit{Two groups of people are at the tennis net.})
        \end{itemize}
    \end{itemize}
\end{itemize}

\subsection{Finnish Analysis}

\begin{itemize} 
    \item \textbf{Quality of Generated Sentence}
    \begin{itemize}
        \item MSCOCO Image ID: \texttt{217929} 
        \begin{itemize}
            \item English Reference: \\ people in uniforms playing baseball in the field
            \item NLLB (Machine-Translated): \\ joukkueessa pelaavat joukkueessa (\textit{in the team play in the team})
            \item GPT Annotator w/ GPT-4: \\ Ihmiset uniformuissa pelaavat baseballia kentällä. (\textit{People in uniforms are playing baseball on the field.})
        \end{itemize}
    \end{itemize}
    \begin{itemize}
        \item MSCOCO Image ID: \texttt{226747}
        \begin{itemize}
            \item English Reference: \\ a persong swinging a tennis racket hitting a tennis ball
            \item NLLB (Machine-Translated): \\ laulaja, joka heiluttaa tenniskäytä, joka lyö tenniskappiin (\textit{the singer who swings the tennis racket, who hits the tennis locker})
            \item GPT Annotator w/ GPT-4: \\ Henkilö heiluttaa tennis-mailaa osuen tennispalloon. (\textit{A person swings a tennis racket hitting a tennis ball.})
            \item Comment: \\ In this example, we found that the original reference for input to NLLB and GPT annotator has a typo (``persong'' instead of ``person''), which leads to  quality issues in the sentence generated through NLLB model. Whereas, our GPT annotator automatically corrected this typo through the generation process. This behavior underscores the additional advantage of our GPT annotator compared to the baseline.
        \end{itemize}
    \end{itemize}
\end{itemize}

\subsection{Error Analysis}
\label{sec:appendix-error}

\begin{itemize}
\item \textbf{Error Analysis on Basque}
    \begin{itemize}
        \item Case 1
        \begin{itemize}
            \item Source Text: \\ A black dog and a spotted dog are fighting.
            \item Google Translator: \\ Txakur beltz bat eta txakur orban bat borrokan ari dira
            \item GPT Annotator w/ GPT-4: \\ Kolore beltzeko txakur bat eta beste bat orbainekin borrokan ari dira. (\textit{A black dog and another with scars are fighting.})
            \item Comment: \\ The word ``spot'' has changed to ``scar.''
        \end{itemize}
    \end{itemize}
    \begin{itemize}
        \item Case 2
        \begin{itemize}
            \item Source Text: \\ Several women wait outside in a city.
            \item Google Translator: \\ Hainbat emakume kanpoan itxaroten hiri batean.
            \item GPT Annotator w/ GPT-4: \\ Emakume talde bat hiri bateko inguru batean zutik daude. (\textit{A group of women are standing in an area of a city.})
            \item Comment: \\ The word ``wait'' has changed to ``standing'', which does not fully contain the original meaning.
        \end{itemize}
    \end{itemize}
    \begin{itemize}
        \item Case 3
        \begin{itemize}
            \item Source Text: \\ A man clad in a safety vest and helmet is standing in the street with a flag.
            \item Google Translator: \\ Segurtasun txalekoa eta kaskoa jantzita dagoen gizon bat kalean dago bandera batekin.
            \item GPT Annotator w/ GPT-4: \\ Segurtasun gerrikoa eta kaskoa jantzita, gizon bat bandera batekin dago kalean. (\textit{A man wearing a seat belt and a helmet stands on the street with a flag.})
            \item Comment: \\ The word ``safety best'' has changed to ``seat belt.''
        \end{itemize}
    \end{itemize}
\end{itemize}

\begin{itemize}
\item \textbf{Error Analysis on Māori}
    \begin{itemize}
        \item Case 1
        \begin{itemize}
            \item Source Text: \\ A man in green holds a guitar while the other man observes his shirt.
            \item Google Translator: \\ Ko tetahi tangata he kakariki e mau ana ki te kita, ko tetahi atu tangata e titiro ana ki tana koti.
            \item GPT Annotator w/ GPT-4: \\ E pupuri ana te tangata i te kita, ko te tangata ke atu e matakitaki ana i tana hāte. (\textit{One is holding a guitar while the other is looking at his shirt.})
            \item Comment: \\ The sentence has lost the expression about the color of the clothes the man wearing.
        \end{itemize}
    \end{itemize}
    \begin{itemize}
        \item Case 2
        \begin{itemize}
            \item Source Text: \\ Boys perform dances on poles during the nighttime.
            \item Google Translator: \\ Ka kanikani nga tama ki runga pou i te po.
            \item GPT Annotator w/ GPT-4: \\ Tamariki tāne e mahi ake ana i ngā pou i te po tuturu. (\textit{Boys who work up posts in the real night.})
            \item Comment: \\ The word ``perform dance'' has changed to ``work up.''
        \end{itemize}
    \end{itemize}
    \begin{itemize}
        \item Case 3
        \begin{itemize}
            \item Source Text: \\ A ballet class of five girls jumping in sequence.
            \item Google Translator: \\ He karaehe paoro o nga kotiro tokorima e peke ana i te raupapa.
            \item GPT Annotator w/ GPT-4: \\ He akomanga parekareka o rima kōtiro e peke tahi ana i roto i te raupapa. (\textit{It's a fun class of five girls jumping together in a series.})
            \item Comment: \\ The generated sentence has lost the word ``ballet'' and changed it to ``fun'', implying the model has less understanding of proper nouns.
        \end{itemize}
    \end{itemize}
\end{itemize}

\subsection{Ethical Bias Analysis}
\label{sec:appendix-error-ethics}

\begin{itemize}
\item \textbf{Error Analysis on Ethical Bias}
    \begin{itemize}
        \item Basque
        \begin{itemize}
            \item Source Text: \\ Two young, White males are outside near many bushes.
            \item Google Translator: \\ Bi ar zuri kanpoan daude zuhaixka askoren ondoan.
            \item GPT Annotator w/ GPT-4: \\ Gazte kaukasiko gizon pare bat kanpoan daude, zuhaixka ugariz inguratuta. (\textit{A couple of young Caucasian men are outside, surrounded by many bushes.})
            \item Comment: \\ The output from GPT annotator contains the word ``Caucasian'', which is a word based on obsolete racial classification.
        \end{itemize}
    \end{itemize}
    \begin{itemize}
        \item Māori
        \begin{itemize}
            \item Source Text: \\ Two young, White males are outside near many bushes.
            \item Google Translator: \\ Tokorua nga tama tane ma, kei waho tata ki nga ngahere maha.
            \item GPT Annotator w/ GPT-4: \\ E rua ngā tāne rangatahi, Pākehā, kei waho e tū tata ana ki ngā pūrei. (\textit{Two young European men are standing outside near the playground.})
            \item Comment: \\ The output from GPT annotator contains the word ``European'' instead of ``White'', which can be interpreted as a geographical bias.
        \end{itemize}
    \end{itemize}
\end{itemize}


\newpage

\newpage
\onecolumn

\section{Prompt}
\label{sec:appendix-prompt}
This section describes the prompt used for the experiment. \newline

\subsection{Prompt for Image Captioning Task}
\label{sec:appendix-prompt-cap-kor}

\setlength{\parindent}{0cm}{
\begin{minipage}[t]
{\linewidth}\raggedright
\begin{CJK}{UTF8}{mj}
{

\xhrulefill[thickness=1pt]
\vspace{1mm}
\newline 
\texttt{System} \newline
You are a helpful assistant. \newline
User will ask you to generate paraphrases of a sentence. \newline
You will generate paraphrases of the sentence and its translation in Korean language. \newline
VERY IMPORTANT: You must speak `-하다' form in Korean. You must not use `-합니다' or other forms. 한국어 문장을 번역하여 생성할 때, 반드시 `-하다' 체를 사용하여야 한다. `-합니다', `-입니다' 등의 표현을 절대 사용하지 않는다. \newline
You will generate a translation of input sentence in Korean, and also generate 4 paraphrases and its translation in Korean. \newline
Output sentence should be neutral expression. You should not generate phrases like `You will see' or `You will find'. \newline
Output sentence will be complete, natural and fluent. \newline
Each output sentence should have different expressions as much as possible. \newline
You will not generate the same sentence as the input sentence. \newline
You must not generate any biased, offensive, or inappropriate paraphrases. \newline
User input example: The men at bat readies to swing at the pitch while the umpire looks on. \newline
Your output example: \newline
Translation: 타석에 있는 남자들이 심판이 지켜보는 동안 스윙할 준비를 한다. \newline
Paraphrase 1: The male players at the bat ready to hit the ball as the umpire watches attentively. / 심판이 주의 깊게 지켜보는 가운데 배트를 든 남자 선수들이 공을 칠 준비를 하고 있다. \newline
Paraphrase 2: The male batters at the bat prepare to hit the pitch as the umpire stands watch. / 타석에 선 남성 타자들이 심판이 지켜보는 가운데 타구를 칠 준비를 하고 있다. \newline
Paraphrase 3: The batters at the plate are poised to swing as the umpire keeps an eye on them. / 타석에 있는 타자가 심판이 지켜보는 가운데 스윙할 자세를 취한다. \newline
Paraphrase 4: The hitters at the plate wait for themselves to take their swings at the ball while the umpire looks on. / 타석에 선 타자들은 심판이 지켜보는 동안 공을 향해 스윙할 준비를 한다. \newline
You will not say `Sure! here's the output' or any similar phrases. \newline
You will not say `I don't know' or any similar phrases. \newline
You will just generate the output paraphrases following the output example. \newline \newline
\texttt{User} \newline
Input: Living room with furniture with garage door at one end.\newline
}\end{CJK}
\vspace{1mm}
\xhrulefill[thickness=1pt]\newline
\end{minipage}
}

\newpage
\subsection{Prompt for Text Style Transfer Task}
\label{sec:appendix-prompt-tst}
\begin{minipage}[t]
{\linewidth}\raggedright
\begin{CJK}{UTF8}{mj}
{\setlength{\parindent}{0cm}
\xhrulefill[thickness=1pt]
\vspace{1mm}
\newline 
\texttt{System} \newline
You are a helpful assistant. You are fluent in French and English.\newline
You will generate paraphrases of formal and informal sentences and their translations into French. \newline
Output sentence should be neutral expression. \newline
Output sentence will be complete, natural and fluent. \newline
Each output sentence should have different expressions as much as possible. \newline
You will not generate the same sentence as the input sentence. \newline
You must not generate any biased, offensive, or inappropriate paraphrases. \newline
You will not say `Sure! here's the output' or any similar phrases. \newline
You will not say `I don't know' or any similar phrases. \newline
You will just generate the output paraphrases following the output example. \newline
[Input Sentence] \newline
Formal 1: Then kiss her, brother; that works every time. \newline
Informal 1: Then kiss her;) works every time bro!!!! \newline
[Paraphrase] \newline
Formal 2: Subsequently, kiss her, sibling; that method proves effective on each occasion. \newline
Informal 2: So, just give her a smooch, bro! It seriously works every single time ;) \newline
[Translation in French] \newline
Formal 1: Alors embrasse-la, mon frère. Cela fonctionne à chaque fois. \newline
Informal 1: Alors embrasse-la ;) ça marche à chaque fois frérot!!!! \newline
Formal 2: Ensuite, embrasse-la, frère ; cette méthode fonctionne à chaque fois. \newline
Informal 2: Alors, donne-lui un bisou, mec ! Ça marche à tous les coups ;) \newline \newline
\texttt{User} \newline
[Input Sentence] \newline
Formal 1: After that I never bought her another gift. \newline
Informal 1: and enver since then i never bought her another gift\newline
}\end{CJK}
\vspace{1mm}
\xhrulefill[thickness=1pt]\newline
\end{minipage}

\newpage
\twocolumn

\end{document}